\documentclass[letterpaper, 10 pt, conference]{ieeeconf}  

\IEEEoverridecommandlockouts                              

\usepackage{amsmath} 
\usepackage{amssymb}  
\usepackage{booktabs}
\usepackage{multirow}
\usepackage{array}
\usepackage{pifont}
\usepackage[table]{xcolor}
\usepackage{threeparttable}
\usepackage{adjustbox}
\usepackage{algorithm}
\usepackage{algpseudocode}
\usepackage{graphicx}
\usepackage{subcaption}
\usepackage{url}
\usepackage{balance}

\title{\LARGE \bf
OpenBelief-Nav: Evidence-Preserving Object Memory for Open-Vocabulary Language-Guided Navigation
}

\author{
Dinh Tuan Nguyen$^{1,*}$, Anh Dao$^{1,*}$, \\
Phuong~Nam Dang$^{1}$, Quan-Dung Pham$^{1}$,
Tuyen P. Le$^{1}$, Truong Nguyen$^{1}$, Quan Nguyen$^{2}$%
\thanks{$^{1}$VinMotion, Inc., Vietnam}
\thanks{$^{2}$University of Southern California, USA}
\thanks{$^{*}$Equal contribution}
}
\begin{document}

\newcommand{\rev}[1]{\textcolor{red}{#1}}
\newcommand{\tbd}{\textcolor{orange}{\textbf{[TBD]}}}
\newcommand{\ours}{OpenBelief\mbox{-}Nav}

\maketitle
\thispagestyle{empty}
\pagestyle{empty}

\begin{abstract}
Open-vocabulary 3D scene graphs provide compact semantic memory for
language-guided navigation, but mapped objects are often exposed through a
single fused feature or committed semantic label. Such commitment can remove
minority yet task-relevant hypotheses from the task-time interface. We present
\ours{}, an evidence-preserving object memory that retains observation-level
phrases, reliability cues, and frame--mask provenance while maintaining
separate aggregate geometric and visual representations. Semantically related
phrases are consolidated into a vocabulary-independent object belief from
which task-specific readouts perform fixed-vocabulary projection or free-form
retrieval. On five ScanNet200 and eight Replica scenes, full-belief projection
achieves mIoU scores of 0.2742 and 0.2912, compared with 0.2393 and 0.2701 for
a matched early-commit readout. Across 78 HM3D--YCB navigation trials,
consensus and early-commit retrieval each achieve 60/78 successes, compared
with 58/78 for belief-weighted retrieval and 55/78 for DualMap. Across 20 Unitree G1 runs organized as 10 matched evaluation
cases, a correction policy permitting at most two verified
candidate attempts improves target-confirmation success from
6/10 to 8/10 relative to top-1-only execution. \textit{Code will be released upon acceptance at} \url{https://openbelief-nav.github.io/}.
\end{abstract}

\begin{figure*}[t]
\centering
\IfFileExists{figures/openbelief_nav.jpg}{%
\includegraphics[width=0.85\textwidth]{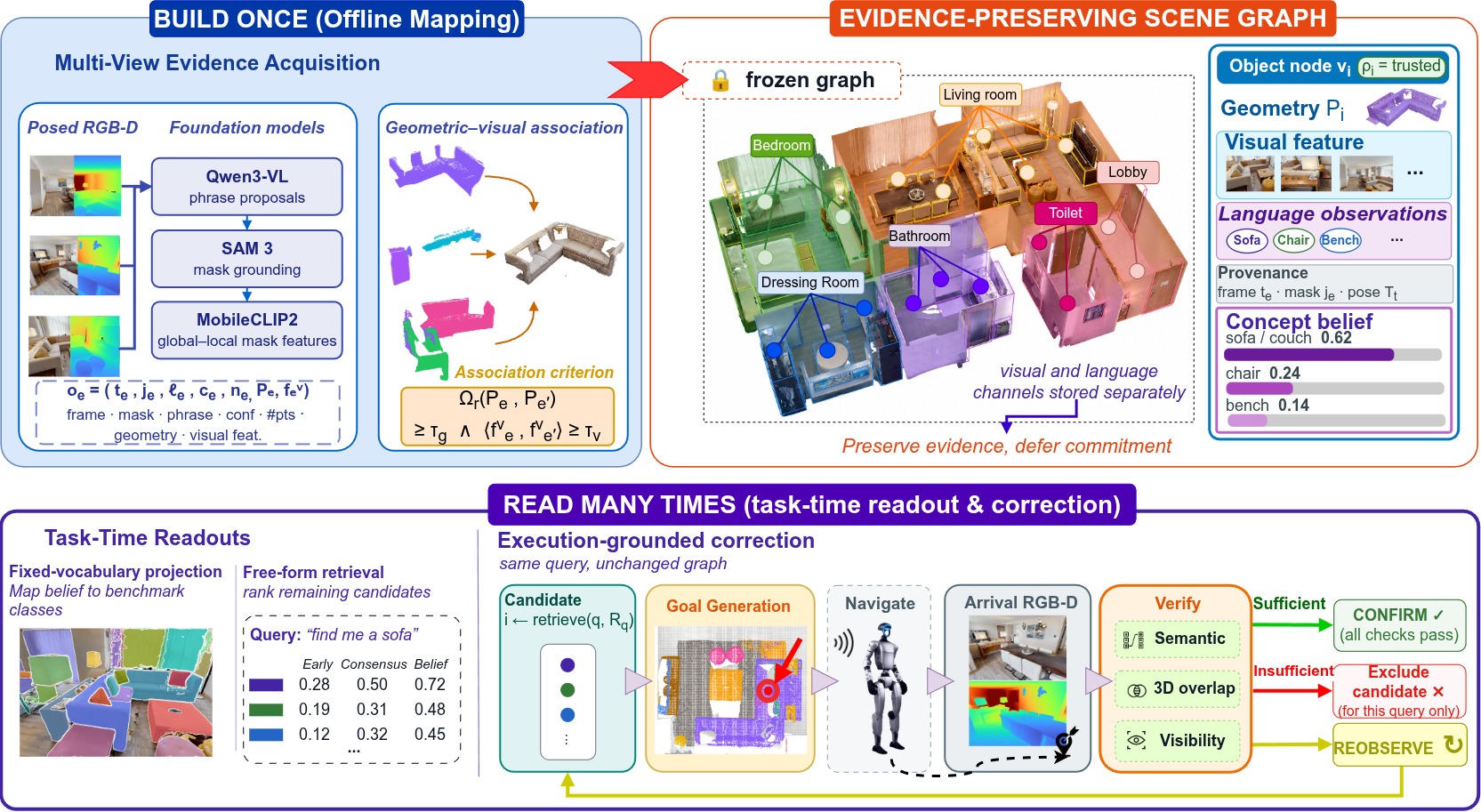}%
}{%
\fbox{%
\parbox[c][0.19\textheight][c]{0.96\textwidth}{%
\centering
}%
}%
}
\caption{Overview of \ours. Mapping retains observation-level phrases,
reliability cues, frame--mask provenance, and separate aggregate geometric and
visual representations. Task-time readouts perform fixed-vocabulary projection
or free-form retrieval, while arrival verification enables query-specific
candidate correction without modifying the persistent graph.}
\label{fig:pipeline}
\end{figure*}

\section{Introduction}

Language-guided navigation requires a robot to ground natural-language concepts in persistent spatial knowledge. Given a request such as ``take me to the fire extinguisher,'' the robot must identify the most plausible mapped instance and produce a reachable navigation goal.

Open-vocabulary 3D scene graphs combine metric geometry with
language-aligned semantics. ConceptGraphs~\cite{gu2024conceptgraphs} builds
object-centric graphs from multi-view detections, HOV-SG~\cite{werby2024hierarchical}
organizes features hierarchically, FSR-VLN~\cite{zhou2025fsr} adds view-level
retrieval, and DualMap~\cite{jiang2025dualmap} supports online target
reselection. Their standard task-time interfaces nevertheless rely primarily
on an aggregate descriptor, selected view, or consensus object state rather
than exposing multiple same-object linguistic hypotheses as a reusable
readout interface.

We identify three consequences of early semantic commitment.

\textbf{First, early commitment can discard recoverable evidence.}
Different views may describe the same object inconsistently because of
occlusion, distance, segmentation noise, or foundation-model uncertainty.
A dominant label may therefore be incorrect even when the correct concept
remains present in the observation history.

\textbf{Second, one semantic readout need not be best for every task.}
Closed-vocabulary segmentation requires projection onto a supplied label set,
whereas navigation retrieval must ground a free-form query. Under the evaluated protocols, full-belief projection achieves the highest
fixed-vocabulary segmentation performance, whereas consensus and
early-commit retrieval obtain the highest navigation success.

\textbf{Third, a top-ranked target remains a hypothesis during execution.} A semantically plausible node may correspond to a distractor or to an instance that cannot be confirmed from the arrival observation. The robot should therefore verify the selected query-to-node binding and, when verification fails, re-ground the unchanged query over the remaining mapped candidates.

We introduce \ours, an open-vocabulary object-memory framework based on delayed
semantic commitment. Each linguistic observation retains its phrase,
reliability cues, and frame--mask provenance, while the associated object node
maintains aggregate geometry and a separate aggregate visual feature.
Multi-view phrases are consolidated into object-specific concept hypotheses
and accumulated as a regularized semantic belief without assigning a
benchmark-specific class during mapping.

At task time, separate readouts perform fixed-vocabulary projection and reliability-aware free-form retrieval. During physical execution, arrival RGB-D observations can confirm, re-observe, or reject a selected candidate. Rejected candidates are excluded only for the current query, and the unchanged instruction is re-grounded over the remaining mapped candidates without modifying the persistent graph. The mechanism therefore corrects the current query-to-node binding rather than performing general dynamic-map repair or absent-target inference.
Figure~\ref{fig:pipeline} summarizes the pipeline.

We evaluate \ours{} on open-vocabulary semantic segmentation and
language-guided navigation using ScanNet200 \cite{dai2017scannet}, Replica \cite{straub2019replica} and HM3D \cite{yadav2023habitat} scenes augmented with YCB
objects \cite{calli2017yale}. The controlled comparisons show task-dependent behavior among semantic
readouts while keeping the stored observations and object geometry fixed
within each evaluation.We additionally evaluate execution-grounded candidate correction
through 20 real-world runs organized as 10 matched evaluation
cases on a Unitree G1 humanoid. Under correct and confusable candidate conditions, correction increases the observed task success from 6/10 (60\%) to 8/10 (80\%).

\textbf{Contributions.} 
\begin{enumerate} 
    \item \textbf{An evidence-preserving object-memory interface.} We introduce an object-centric representation that retains per-observation language phrases, reliability cues, and frame--mask provenance while maintaining a separate aggregate visual feature, enabling semantic decisions to be recomputed at task time. 
    \item \textbf{Task-specific readouts over a shared frozen graph.}
We instantiate full-belief fixed-vocabulary projection together with
consensus, early-commit, and belief-weighted free-form retrieval as separate readouts over the same stored object evidence. Within each controlled comparison, the variants use identical observations and object geometry.
    \item \textbf{Query-specific execution correction.} We formulate a query-preserving verification loop that excludes a candidate failing arrival verification only for the current query and re-grounds the unchanged request over the remaining mapped candidates without modifying the persistent graph. The loop terminates when a candidate is confirmed, no untried mapped candidate remains, or the candidate-attempt budget is reached. We evaluate the mechanism in a 20-run pilot study comprising 10 matched evaluation cases on a Unitree G1 humanoid.
    
\end{enumerate}

\section{Related Work}

\subsection{Open-Vocabulary Semantic Mapping}

Open-vocabulary mapping attaches language-aligned features to geometric map elements.
ConceptFusion~\cite{jatavallabhula2023conceptfusion} and
VLMaps~\cite{huang2023visual} embed vision-language features densely into
points and grids; ConceptGraphs~\cite{gu2024conceptgraphs} lifts this to
object-centric 3D scene graphs by incrementally fusing per-detection CLIP
features into a single running-average embedding per object.
HOV-SG~\cite{werby2024hierarchical} builds a hierarchical floor--room--object
graph for large environments.
OGScene3D~\cite{zhu2026ogscene3d} incrementally maintains semantic 3D Gaussians and scene-graph relations using confidence-aware optimization.
ThinkGraphs~\cite{bickici2026think} maintains incremental object tracks and asynchronously enriches them with VLM-derived labels and descriptions.
These methods demonstrate the value of structured semantic memory, but their downstream interfaces still rely on a committed or consensus object representation.
Our focus is complementary: we retain object-specific observation-level
linguistic evidence and expose it to multiple task-time readouts while
maintaining aggregate geometric and visual representations.

Classical semantic mapping has long maintained fixed-vocabulary class distributions, for example in SemanticFusion~\cite{mccormac2017semanticfusion}.
Our representation differs in that the concept set is not fixed during map construction.
Each node accumulates object-specific phrases, which are projected onto a consumer vocabulary only at read time.
We use Dirichlet-inspired pseudo-count smoothing and adopt ``evidence'' in the broad subjective-logic sense~\cite{jsang2018subjective,sensoy2018evidential}; we do not claim learned evidential calibration.

\subsection{Scene Graphs for Language Navigation}

FSR-VLN~\cite{zhou2025fsr} performs efficient hierarchical retrieval and then applies slower VLM reasoning.
SysNav~\cite{zhu2026sysnav} separates semantic reasoning, planning, and embodiment-specific control for real-world cross-embodiment navigation.
Our method does not propose a new low-level navigation architecture.
Instead, it improves the object-memory and target-grounding interface supplied to a conventional planner.

\subsection{Dynamic Mapping and Target Correction}

DualMap combines an abstract global map with online local mapping and supports target reselection when a queried object changes location~\cite{jiang2025dualmap}. OpenIN~\cite{tang2025openin} and
DOVSG~\cite{yan2025dynamic} update carrier-relation or dynamic scene graphs when
objects relocate.
ThinkGraphs repairs fragmented tracks during graph construction using asynchronous VLM agents~\cite{bickici2026think}.
Our execution correction has a narrower objective. It does not update the persistent graph, discover a newly introduced object, infer that a queried object is globally absent, or claim general relocation recovery. Instead, an arrival RGB-D observation tests the current query-to-node hypothesis. A candidate that fails verification is excluded only for that query, and the unchanged instruction is re-grounded over the remaining mapped candidates. The procedure ends in failure when no candidate is confirmed within the candidate-attempt budget or when no untried mapped candidate remains.

Table~\ref{tab:positioning} positions \ours{} against the most closely
related systems along the axes developed above.

\definecolor{ourrow}{RGB}{210, 235, 245}       
\definecolor{cmarkcolor}{RGB}{0, 128, 96}      
\newcommand{\cmark}{\textcolor{cmarkcolor}{\ding{51}}}
\newcommand{\xmark}{\ding{55}}
\begin{table}[t]
\centering
\caption{Comparison with related open-vocabulary scene-graph systems.
``Evidence @ query'' denotes multiple same-object linguistic hypotheses
available to task-time readouts. \cmark: supported; $\triangle$: supported
through a different formulation; \xmark: unsupported.}
\label{tab:positioning}
\setlength{\tabcolsep}{2.6pt}
\resizebox{\columnwidth}{!}{%
\begin{tabular}{lcccc}
\toprule
System & Open-vocab & \shortstack{Evidence\\@ query}
& \shortstack{Exec.-grounded\\correction}
& \shortstack{Training-\\free} \\
\midrule
ConceptGraphs~\cite{gu2024conceptgraphs} & \cmark & \xmark & \xmark & \cmark \\
HOV-SG~\cite{werby2024hierarchical}      & \cmark & \xmark & \xmark & \cmark \\
FSR-VLN~\cite{zhou2025fsr}               & \cmark & \xmark & \xmark & \cmark \\
DualMap~\cite{jiang2025dualmap}          & \cmark & \xmark & $\triangle$ & \cmark \\
\rowcolor{ourrow}
\ours{} (ours)                           & \cmark & \cmark & \cmark & \cmark \\
\bottomrule
\end{tabular}%
}
\end{table}

\section{Problem Formulation}

\label{sec:problem}
We are given a stream of posed RGB-D observations
\(\mathcal{D}=\{(I_t,D_t,T_t)\}_{t=1}^{N}\) with camera-to-world poses
\(T_t\in SE(3)\), and must build a persistent spatial memory that answers
free-form language queries with reachable navigation goals. We organize the
environment as a hierarchical scene graph over floor, room, view, and object
nodes; our contribution concerns the object nodes and the interface they expose
to downstream tasks.

Conventional maps often compress an object's observations into one label or
fused embedding during construction. A task-relevant minority hypothesis
discarded by this commitment cannot be recovered at read time without
rerunning perception. We instead retain observation-level linguistic evidence
and an independent aggregate visual feature, deferring semantic decisions to
task-time readouts. The resulting object state \(\mathcal B_i\) is defined in
Section~\ref{sec:overview}.

Given a query \(q\), the system produces a per-object score \(S_i(q)\) and selects the highest-scoring mapped candidate that has not previously been excluded for that query. During deployment, the selected target is treated as a hypothesis rather than a fixed semantic fact. At correction attempt \(k\), the robot acquires an arrival observation \(z^{(k)}\) and verifies the selected candidate \(i^{(k)}\). If verification fails, the candidate is added to a query-specific exclusion set, \begin{equation} \mathcal{R}^{(k+1)}_q = \mathcal{R}^{(k)}_q\cup\{i^{(k)}\}, \qquad q^{(k+1)}=q. \label{eq:problem-update} \end{equation} The unchanged query is then re-grounded over the remaining mapped candidates. The persistent graph is not modified. Execution terminates when a candidate is confirmed, no untried mapped candidate remains, or the candidate-attempt budget is reached. The design question addressed by this paper is where semantic commitment should occur across map construction, task-time readout, and physical
verification.

\section{Method}
\subsection{Overview: object nodes as evidence, not decisions}
\label{sec:overview}
In a pipeline that assigns each object a single label and embedding at
construction time, a task-relevant category can remain present in the
accumulated multi-view phrases yet be absent from the committed readout. Early commitment therefore does not merely risk perceptual error: it can
remove already acquired semantic hypotheses from the task-time object state.
Our
response is to defer commitment. An object node stores the evidence together
with a cheap summary of it, and every semantic decision is computed downstream
from that evidence:
\begin{equation}
\mathcal{B}_i=
\left(P_i,f_i^v,\mathcal{O}_i,\mathbf{V}_i,\mathbf{p}_i,f_i^t,\rho_i\right),
\label{eq:node-state}
\end{equation}
where \(P_i\) is the point cloud, \(f_i^v\) is an aggregate visual
embedding, \(\mathcal{O}_i\) is the list of language observations,
\(\mathbf{V}_i=\{V_{ik}\}_{k=1}^{K_i}\) is the concept-evidence vector,
\(\mathbf{p}_i=\{p_{ik}\}_{k=1}^{K_i}\) is its normalized relative-support
distribution, \(f_i^t\) is the consensus text feature defined in
Eq.~\eqref{eq:consensus-feature}, and \(\rho_i\) is the trust status defined
in Sec.~\ref{sec:belief}. The representation obeys a separation
invariant: the visual channel \(f_i^v\) and the language channels
\((\mathcal{O}_i,\mathbf{V}_i,\mathbf{p}_i,f_i^t)\) are maintained
independently and are never destructively fused during mapping; all cross-modal
fusion is deferred to the task-time readouts of Section~\ref{sec:readouts}. We
are precise about what is preserved: the language evidence is retained per
observation, whereas the visual embedding is an aggregate. The remainder of the method consists of four
stages—evidence acquisition, geometric–visual association, belief
construction, and task-time readout with execution-time verification—each of which defers task-specific commitment while
preserving the evidence required by downstream readouts.

\subsection{Multimodal evidence acquisition}
\label{sec:acquisition}
For each frame \(I_t\), a vision--language model proposes a set of object
phrases and a promptable segmentation model grounds each phrase as one or more
masks, so that every mask retains the phrase that produced it together with its
frame and mask provenance. Following FSR-VLN~\cite{zhou2025fsr}, we compute each mask's appearance feature using global--local CLIP aggregation and back-project its valid depth pixels into the world frame using \(T_t\). The mask is stored as a single observation
\begin{equation}
o_e=(t_e,j_e,\ell_e,c_e,n_e,P_e,f_e^v),
\label{eq:observation}
\end{equation}
where the entries denote the source frame, mask index, phrase, segmentation
confidence, valid-point count, back-projected point cloud, and normalized
visual feature, respectively. The unit of preserved semantic evidence is the
observation, not the final object. Multiple observations assigned to one object
may therefore retain different or contradictory phrases. This disagreement is
not removed during mapping; it becomes the input to the belief construction.

\subsection{Geometric--visual object association}
\label{sec:association}
Per-frame observations must be associated into object instances before semantic
belief is inferred. Association cannot rely on generated prompts,
predicted names, or semantic agreement, since doing so would make instance
formation depend on the uncertain evidence that the subsequent belief model is
intended to resolve. We instead adopt conservative geometry-based association, using visual similarity only to reject inconsistent matches:
\begin{equation}
o_e\sim o_{e'}.
\Longleftrightarrow
\Omega_r(P_e,P_{e'})\ge \tau_g
\ \wedge
\left\langle f_e^v,f_{e'}^v\right\rangle\ge \tau_v,
\label{eq:association}
\end{equation}
where \(P_e\) and \(f_e^v\) denote the spatial support and normalized visual feature of observation \(o_e\) and \(\Omega_r\) is the
radius-based point-cloud overlap function used by the mapping pipeline. Object instances are formed as connected components of this relation. We favor high-precision association because false merges
irreversibly mix evidence from distinct objects, whereas missed
associations remain as separate candidates. Downstream grounding
may select among these candidates but does not repair their
object identities.

\subsection{Evidence-preserving semantic belief}
\label{sec:belief}
Given a node's observations, we summarize the language evidence as a
smoothed relative-support distribution over its retained concept hypotheses.
Pseudo-count smoothing regularizes competition among retained hypotheses,
while an observation-count trust status separately limits reliance on sparse
language evidence.

\emph{Reliability:} VLM phrase proposals carry no usable per-object confidence,
so we weight each observation only by quantities we do possess---its
segmentation confidence and the square root of its 3D support:
\begin{equation}
w_e=c_e\sqrt{n_e}.
\label{eq:reliability-weight}
\end{equation}

\emph{Synonym-aware concept consolidation:}
Multi-view observations may describe the same object using semantically equivalent phrases, causing its evidence to be fragmented across lexical variants. After geometric association, we embed the observed phrases in a shared text space and group semantically similar expressions into object-level concept hypotheses. The support for concept k of object i is accumulated as
\begin{equation}
V_{ik} =
\sum_{o_e \in \mathcal{O}_i:\,\ell_e \in \mathcal{C}_{ik}} w_e,
\label{eq:concept-vote}
\end{equation}
where \(\mathcal{C}_{ik}\) denotes the phrase cluster associated with that concept. This consolidation reduces synonym-induced vote splitting while retaining semantically distinct hypotheses as competing alternatives.

\emph{Low-evidence regularization:}
For \(K_i>0\), symmetric pseudo-count smoothing gives
\begin{equation}
p_{ik}=
\frac{V_{ik}+\alpha}
{\sum_{j=1}^{K_i}V_{ij}+\alpha K_i},
\qquad \alpha>0.
\label{eq:dirichlet-belief}
\end{equation}
This preserves nonzero relative support for retained competing hypotheses but
is not a calibrated uncertainty estimate. We separately set $\rho_i=\mathrm{degraded}$ when
$|\mathcal{O}_i|<m_{\min}$ and $\rho_i=\mathrm{trusted}$
otherwise. During free-form retrieval, degraded nodes place less
weight on the language branch.

\emph{Consensus text feature:} For nodes with \(K_i>0\), we additionally maintain a consensus text feature, the belief-weighted combination of
its consolidated concept embeddings,
\begin{equation}
f_i^t=\operatorname{normalize}\!\Big(\sum_{k=1}^{K_i} p_{ik}\,
g_{ik}\Big),
\label{eq:consensus-feature}
\end{equation}
where \(g_{ik}\) is the normalized text embedding of concept
\(\mathcal{C}_{ik}\). This is a summary for scoring only; the underlying
belief \(\mathbf{p}_i\) remains available to every readout.

\subsection{Task-time semantic readouts}
\label{sec:readouts}
Because a node stores evidence rather than a decision, ``what is this object?''
is answered differently for different tasks over the same frozen graph.
We instantiate fixed-vocabulary projection and free-form retrieval and
compare alternative semantic readouts over identical stored observations and
object geometry.

\emph{Reliability-aware free-form retrieval:}
Given a normalized query embedding \(f_q\), the three evaluated language
readouts are
\begin{align}
H_i^{\mathrm{con}}(q)
&=\left\langle f_i^t,f_q\right\rangle,
\\
H_i^{\mathrm{early}}(q)
&=\left\langle g_{ik_i^\star},f_q\right\rangle,
\qquad
k_i^\star=\arg\max_k V_{ik},
\\
H_i^{\mathrm{bel}}(q)
&=\sum_{k=1}^{K_i}p_{ik}
\left\langle g_{ik},f_q\right\rangle .
\label{eq:retrieval-heads}
\end{align}
For a selected readout \(r\in
\{\mathrm{con},\mathrm{early},\mathrm{bel}\}\), the object score is
\begin{equation}
S_i^{(r)}(q)=
\lambda_i\left\langle f_i^v,f_q\right\rangle+
(1-\lambda_i)H_i^{(r)}(q),
\label{eq:retrieval-score}
\end{equation}
where
\begin{equation}
\lambda_i=
\begin{cases}
1.0, & K_i=0,\\
0.7, & \rho_i=\mathrm{degraded},\\
0.2, & \rho_i=\mathrm{trusted}.
\end{cases}
\label{eq:reliability-fusion}
\end{equation}
Thus, nodes without linguistic evidence are scored visually, degraded nodes
lean toward the visual branch, and trusted nodes place greater weight on the
selected language readout.

Let \(\mathcal V_o\) denote the mapped object nodes and
\(\mathcal A_q=\mathcal V_o\setminus\mathcal R_q\) the candidates not excluded
for query \(q\). Retrieval is defined as \begin{equation} \operatorname{retrieve}(q,\mathcal{R}_q)= \begin{cases} \mathrm{NONE}, & \mathcal{A}_q=\varnothing,\\ \displaystyle\arg\max_{i\in\mathcal{A}_q}S_i^{(r)}(q), & \text{otherwise}. \end{cases} \label{eq:retrieval-rule} \end{equation} The sentinel \(\mathrm{NONE}\) indicates only that no untried mapped candidate remains under the current query-specific exclusion state. It is not an absent-query prediction and does not imply that the queried object is absent from the environment.

\emph{Delayed semantic commitment for fixed-vocabulary readout:}
Graph construction operates over free-form concepts, whereas benchmark segmentation requires classification over a closed vocabulary available only at evaluation time. We therefore avoid early benchmark-label assignment and retain for each object (i) a compact belief \(\mathbf{p}_i=\{p_{ik}\}_{k=1}^{K_i}\) over its consolidated concepts. Although not lossless, this representation remains independent of the downstream vocabulary and preserves plausible secondary hypotheses discarded by an early top-1 decision.

Once the target vocabulary is known, visual evidence refines the belief only within its retained semantic support:
\begin{equation}
\widetilde p_{ik}=
\frac{p_{ik}^{\gamma}
\exp\left(\left\langle g_{ik},f_i^v\right\rangle/T_{\mathrm{vis}}\right)}
{\sum_{j=1}^{K_i}p_{ij}^{\gamma}
\exp\left(\left\langle g_{ij},f_i^v\right\rangle/T_{\mathrm{vis}}\right)},
\label{eq:belief-refinement}
\end{equation}
where \(g_{ik}\) is the text embedding of concept \(\mathcal{C}_{ik}\), \(0<\gamma\le1\) tempers an over-concentrated support distribution, and \(T_{\mathrm{vis}}\) controls visual re-ranking. This support-preserving refinement cannot introduce concepts outside the retained belief, but may promote a visually consistent retained secondary hypothesis.

For benchmark class \(c\), define the direct visual prediction
\begin{equation}
v_i(c)=
\operatorname{softmax}_{c}\!\left(
\frac{\left\langle f_i^v,t_c\right\rangle}{T_{\mathrm{obj}}}
\right),
\end{equation}
where \(t_c\) is the normalized text embedding of class \(c\), and define
the concept-to-class affinity
\begin{equation}
W_{ikc}=
\operatorname{softmax}_{c}\!\left(
\frac{\langle g_{ik},t_c\rangle}{T_{\mathrm{proj}}}
\right).
\end{equation}
The full-belief class distribution is
\begin{equation}
\widehat p_i(c)=
\begin{cases}
v_i(c), & K_i=0,\\[1mm]
\displaystyle
(1-\lambda_{\mathrm{vis}})
\sum_{k=1}^{K_i}\widetilde p_{ik}W_{ikc}
+\lambda_{\mathrm{vis}}v_i(c), & K_i>0.
\end{cases}
\label{eq:fixed-vocab-readout}
\end{equation}
The predicted benchmark label is
\(\widehat y_i=\arg\max_c\widehat p_i(c)\).

For the controlled readout comparison, \emph{visual-only} uses \(v_i(c)\)
directly. \emph{Consensus} projects the consensus feature as
\begin{equation}
\widehat p_i^{\mathrm{con}}(c)=
\begin{cases}
v_i(c), & K_i=0,\\[1mm]
\displaystyle
\operatorname{softmax}_{c}\!\left(
\frac{\langle f_i^t,t_c\rangle}{T_{\mathrm{proj}}}
\right), & K_i>0.
\end{cases}
\label{eq:consensus-projection}
\end{equation}
\emph{Early-commit} replaces \(\mathbf p_i\) with a one-hot distribution at
\(k_i^\star=\arg\max_k V_{ik}\) before applying the same
benchmark-conditioned projection as Eq.~\eqref{eq:fixed-vocab-readout}.
\emph{Full-belief projection} retains all hypotheses and uses
Eq.~\eqref{eq:fixed-vocab-readout}. Benchmark-specific projection is used
only for semantic-segmentation evaluation and is not stored in the graph.

\subsection{Execution-Grounded Candidate Correction} \label{sec:execution-correction} A retrieved object node is treated as a hypothesis rather than an immediately confirmed target. After navigating to the candidate goal, the robot acquires a local RGB-D observation and applies three ordered checks. First, it checks whether enough of the stored candidate geometry should be visible from the arrival viewpoint. If visibility is below \(\tau_\nu\), the robot performs one bounded re-observation. Second, it checks whether the strongest local query--mask similarity exceeds \(\tau_q\). Third, it checks whether the query-matching local mask geometrically overlaps the stored candidate by at least \(\tau_{\mathrm{arr}}\). The candidate is confirmed only when the visibility, semantic, and geometric checks are satisfied. A candidate that remains unconfirmed after the verification procedure,
including bounded re-observation when required, is excluded only for the
current query. The persistent graph is not modified, and the original query is re-grounded over the remaining mapped candidates. Consequently, the mechanism corrects the current query-to-node binding but does not infer that the queried object is absent, perform general dynamic-map repair, or recover an object at a location absent from the stored graph. The correction procedure assumes that the navigation stack reaches the candidate observation viewpoint. Failures of the underlying navigation stack are not interpreted as semantic evidence against the selected candidate. \begin{algorithm}[t] \small \caption{Query-preserving execution correction} \label{alg:correction} \begin{algorithmic}[1] \Require Query \(q\), frozen graph \(\mathcal G\), attempt budget \(K_{\max}\) \Ensure Outcome in \(\{\mathrm{CONFIRM},\mathrm{FAIL}\}\) \State \(\mathcal R_q\gets\varnothing\) \For{\(k=1,\ldots,K_{\max}\)} \State \(i\gets\operatorname{retrieve}(q,\mathcal R_q)\) \If{\(i=\mathrm{NONE}\)} \State \textbf{return} \(\mathrm{FAIL}\) \EndIf \State Navigate to the candidate goal \(g_i^\star\) \State Acquire an arrival RGB-D observation and evaluate candidate \(i\) \If{the candidate has insufficient visibility} \State Execute one bounded re-observation and re-evaluate candidate \(i\) \EndIf \If{the candidate is confirmed} \State \textbf{return} \(\mathrm{CONFIRM}\) \EndIf \State \(\mathcal R_q\gets\mathcal R_q\cup\{i\}\) \EndFor \State \textbf{return} \(\mathrm{FAIL}\) \end{algorithmic} \end{algorithm}

\subsection{Navigation Interface}

The object's point cloud is projected onto the traversability map. The planner
selects the nearest collision-free boundary pose with expected visibility and
executes it using a conventional global/local navigation stack. Embodiment
control is outside our contribution; \ours{} supplies the object-level semantic
target.
\section{Experiments}

\label{sec}
\subsection{Experimental Setup}
\label{sec:exp_setup}

We evaluate two complementary uses of the same object representation:
fixed-vocabulary 3D semantic segmentation and language-guided object
navigation.

\paragraph{Open-vocabulary semantic segmentation}
Following HOV-SG and DualMap, we evaluate on five ScanNet200
scenes: \texttt{scene0011\_00}, \texttt{scene0050\_00},
\texttt{scene0231\_00}, \texttt{scene0378\_00}, and
\texttt{scene0518\_00}. We additionally evaluate on eight Replica
scenes, \texttt{office0--office4} and \texttt{room0--room2}.
All OpenBelief-Nav variants share identical observations and object
geometry and differ only in their task-time semantic readout.
\emph{Visual-only} uses the direct object visual prediction \(v_i(c)\). \emph{Consensus} projects the
belief-weighted consensus feature using Eq.~\eqref{eq:consensus-projection}.
\emph{Early-commit} retains only the highest-support concept before applying
the same benchmark-conditioned projection. \emph{Full-belief projection}
retains all concept hypotheses and delays projection to the benchmark
vocabulary until evaluation. We follow the
point-wise evaluation protocol of HOV-SG and DualMap and report mean
intersection-over-union (mIoU), frequency-weighted IoU (F-mIoU), and
mean accuracy (mAcc).

\paragraph{Language-guided navigation}
We evaluate in Habitat-Sim \cite{puig2024habitat} on HM3D scenes \texttt{00829},
\texttt{00848}, and \texttt{00880}, following the static navigation
protocol of DualMap with inserted YCB objects. For each query, the
system retrieves an object node and navigates to its mapped location.
A trial succeeds when the final agent position is within
$1\,\mathrm{m}$ of the target, and performance is reported as
SR@1m. The evaluation contains 78 trials per method. We compare ConceptGraphs, HOV-SG, and DualMap as reference systems against
the consensus, early-commit, and belief-weighted retrieval readouts of
OpenBelief-Nav. The controlled claims among OpenBelief-Nav variants are based
on identical graph geometry, observations, and navigation stacks; these
variants differ only in semantic target selection. 

\paragraph{Implementation details}
Phrase proposals are generated by Qwen3-VL-4B-Instruct~\cite{bai2025qwen3},
grounded into object masks using SAM~3~\cite{carion2025sam}, and encoded with
MobileCLIP2-S4~\cite{faghri2025mobileclip2} for both visual and textual features.
Mapping is performed offline on a workstation equipped with one
NVIDIA RTX~5090 GPU (32\,GB), while task-time readouts operate online. Unless stated otherwise,
\((\tau_g,\tau_v,\alpha,m_{\min},\gamma,T_{\mathrm{vis}},
T_{\mathrm{proj}},T_{\mathrm{obj}},\lambda_{\mathrm{vis}})
=(0.75,0.85,0.50,3,0.50,0.10,0.02,0.02,0.25)\), and
\((\tau_\nu,\tau_q,\tau_{\mathrm{arr}},K_{\max})
=(0.25,0.50,0.20,2)\).

\subsection{Open-Vocabulary Semantic Segmentation}
\label{sec:semseg_results}

Table~\ref{tab:semseg_results} reports the main results. Among matched
OpenBelief-Nav readouts, full-belief projection achieves the highest mIoU on
ScanNet200 and Replica, improving over early-commit projection from 0.2393 to
0.2742 and from 0.2701 to 0.2912, respectively. It also obtains the highest
reported mAcc on both datasets and the highest F-mIoU on ScanNet200, while
closely matching DualMap on Replica F-mIoU (0.5204 versus 0.5207). These
controlled comparisons support retaining secondary hypotheses until the
downstream vocabulary is known.

\definecolor{bestcell}{RGB}{186, 230, 201}    
\definecolor{secondcell}{RGB}{255, 224, 178}  
\newcommand{\best}[1]{\cellcolor{bestcell}\textbf{#1}}
\newcommand{\second}[1]{\cellcolor{secondcell}#1}

\begin{table}[t]
\centering
\caption{Open-vocabulary 3D semantic segmentation. OpenBelief-Nav variants
share the same graph. Higher is better; best and second-best results are shown
in bold green and amber.}
\label{tab:semseg_results}
\small
\setlength{\tabcolsep}{3.5pt}
\begin{tabular}{llccc}
\toprule
Dataset & Method
& mIoU $\uparrow$
& F-mIoU $\uparrow$
& mAcc $\uparrow$ \\
\midrule
\multirow{5}{*}{Replica}
& ConceptGraphs    & 0.1501 & 0.3858 & 0.3559 \\
& HOV-SG           & 0.2050 & 0.4846 & 0.3835 \\
& DualMap          & 0.2538 & \best{0.5207} & 0.4024 \\
& Ours (Consensus) & \second{0.2831} & 0.4647 & \second{0.4774} \\
& Ours (Full belief) & \best{0.2912} & \second{0.5204} & \best{0.4822} \\
\midrule
\multirow{5}{*}{ScanNet200}
& ConceptGraphs    & 0.0882 & 0.3077 & 0.3538 \\
& HOV-SG           & 0.1333 & 0.3381 & 0.3714 \\
& DualMap          & 0.1604 & 0.3288 & 0.3794 \\
& Ours (Consensus) & \second{0.2107} & \second{0.4887} & \second{0.4580} \\
& Ours (Full belief) & \best{0.2742} & \best{0.5276} & \best{0.4685} \\
\bottomrule
\end{tabular}
\end{table}

\subsection{Language-Guided Navigation in Simulation}
\label{sec:navigation_results}

Table~\ref{tab:navigation_main} reports SR@1m under the HM3D--YCB
protocol. Consensus and early-commit obtain the highest average score among
the evaluated readouts, 76.9\%, while the belief-weighted readout obtains 74.4\%.
The corresponding DualMap, ConceptGraphs, and HOV-SG reference scores are
70.5\%, 61.5\%, and 52.6\%, respectively. No internal readout dominates every
scene: early-commit is strongest on \texttt{00829}, consensus is strongest on
\texttt{00848}, and the three text-based variants tie on \texttt{00880}.
Together with the segmentation results, this variation motivates
task-specific readouts rather than treating one committed representation as
uniformly optimal across the evaluated tasks.



\begin{table}[t]
\centering
\caption{HM3D--YCB navigation SR@1m (\%); 78 trials per method. Best and
second-best results are shown in bold green and amber.}
\label{tab:navigation_main}
\setlength{\tabcolsep}{4.0pt}
\resizebox{\columnwidth}{!}{%
\begin{tabular}{lcccc}
\toprule
Method & \texttt{00829} & \texttt{00848} & \texttt{00880} & Avg. SR@1m $\uparrow$ \\
\midrule
ConceptGraphs~\cite{gu2024conceptgraphs} & 69.2 & 53.8 & 61.5 & 61.5 \\
HOV-SG~\cite{werby2024hierarchical}      & 53.8 & 46.2 & 57.7 & 52.6 \\
DualMap~\cite{jiang2025dualmap}          & 73.1 & 69.2 & \second{69.2} & 70.5 \\
\midrule
Consensus-only & \second{80.8} & \best{76.9} & \best{73.1} & \best{76.9} \\
Early-commit   & \best{84.6}  & \second{73.1} & \best{73.1} & \best{76.9} \\
Belief-weighted & 76.9         & \second{73.1} & \best{73.1} & \second{74.4} \\
\bottomrule
\end{tabular}}
\end{table}

\balance

\begin{figure*}[!t]
    \centering
    \begin{subfigure}[t]{0.4\textwidth}
        \centering
        \includegraphics[width=\linewidth]{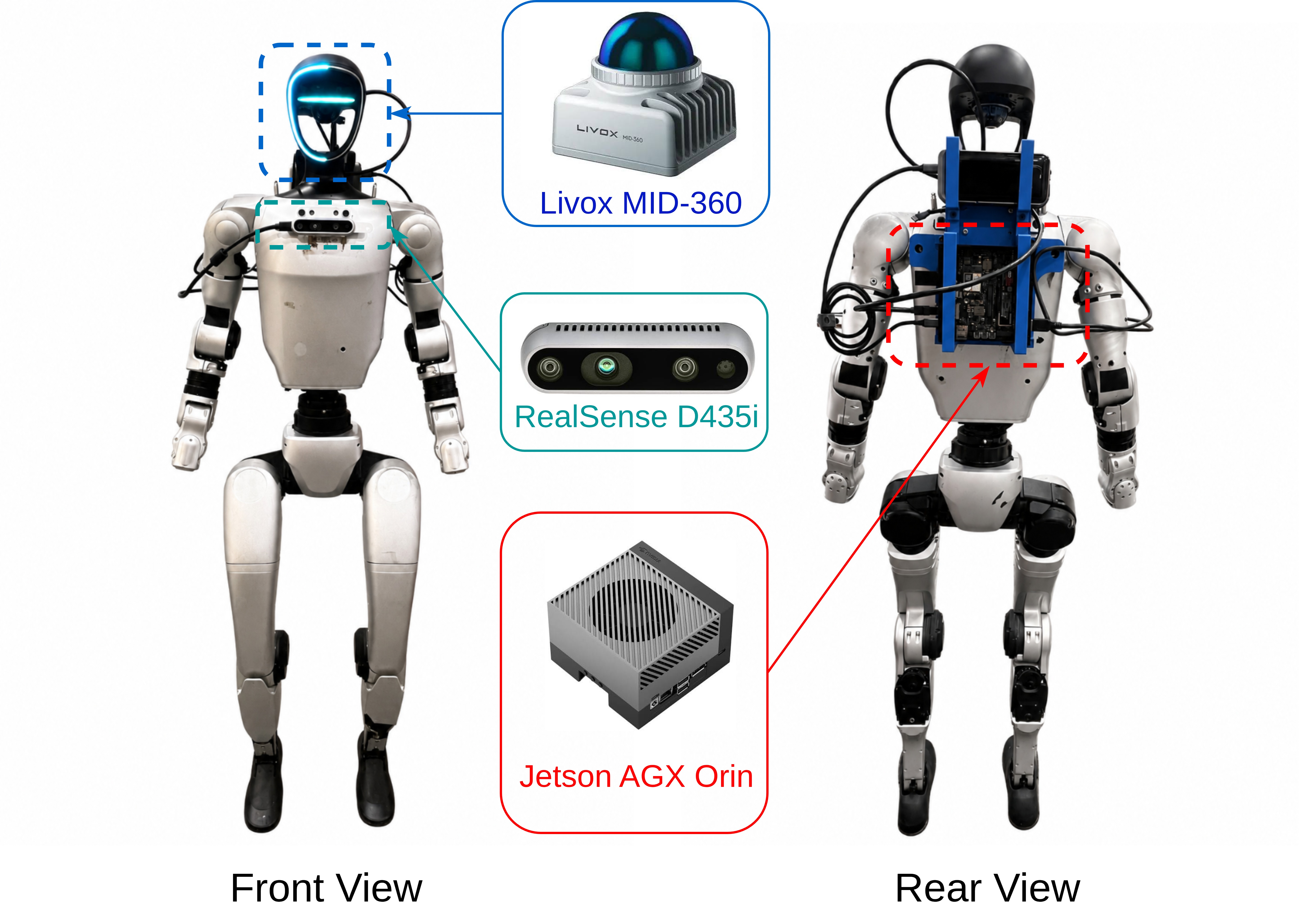}
        \caption{Unitree G1 platform and hardware configuration.}
        \label{fig:realworld_platform}
    \end{subfigure}%
    \hspace{0.02\textwidth}%
    \begin{subfigure}[t]{0.4\textwidth}
        \centering
        \includegraphics[width=\linewidth]{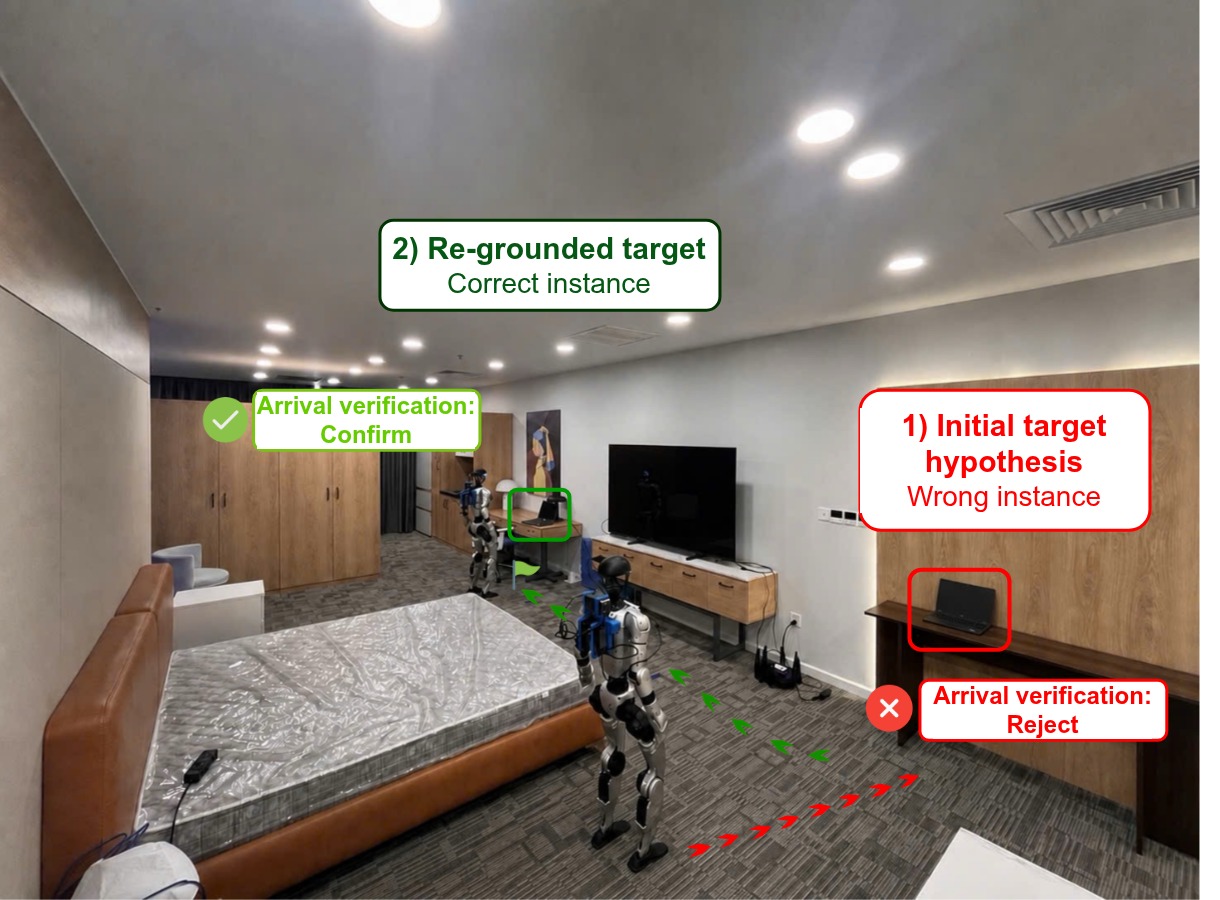}
        \caption{Execution-grounded candidate correction.}
        \label{fig:realworld_correction}
    \end{subfigure}
    \caption{Real-world platform and deployment procedure.}
    \label{fig:realworld_overview}
\end{figure*}

\subsection{Real-World Language-Guided Navigation} \label{sec:navigation_realworld} 

We deploy the proposed pipeline on a Unitree G1 humanoid equipped with an Intel RealSense D435i RGB-D camera and a Livox MID-360 LiDAR (Fig.~\ref{fig:realworld_platform}). An external workstation with an NVIDIA GeForce RTX~5090 GPU and Intel Core Ultra~9 285K CPU performs semantic mapping, open-vocabulary retrieval, and language-conditioned reasoning. An onboard NVIDIA Jetson AGX Orin runs ROS~2 \cite{doi:10.1126/scirobotics.abm6074} and Nav2 \cite{macenski2020marathon2} and interfaces with the low-level robot controller. The two computers communicate through Zenoh, keeping compute-intensive semantic reasoning offboard while preserving onboard motion planning and control. 

Each free-form query is grounded to a target pose using consensus retrieval.
After navigation, arrival RGB-D observations verify the candidate as described
in Section~\ref{sec:execution-correction}. Full correction permits at most
\(K_{\max}=2\) verified candidate attempts.

We evaluate five paired episodes with a correct initial candidate and five
with a competing distractor, using the same graph, query, configuration, and
start pose for top-1-only and full-correction execution. Table~\ref{tab:realworld_main}
shows that success increases from 6/10 to 8/10; under the confusable condition,
it increases from 1/5 to 3/5. Full correction visits 1.40 candidates per
episode on average.

\textbf{Scope.}
This pilot evaluates recovery when the intended target remains mapped; it does
not evaluate absent-query rejection, removed-target detection, general map
repair, or broad real-world generalization.


\begin{table}[t]
\centering
\caption{Matched real-world target-confirmation results.
Each matched case is run once per method using the same frozen
graph, query, object configuration, and reset start pose. The
evaluation therefore contains 20 robot runs in total.}
\label{tab:realworld_main}
\setlength{\tabcolsep}{3.2pt}
\resizebox{\columnwidth}{!}{%
\begin{tabular}{lcccc}
\toprule
Condition
& \shortstack{Runs per\\method}
& \shortstack{Top-1 only\\Confirmation SR}
& \shortstack{Full correction\\Confirmation SR}
& \shortstack{Full correction\\mean visits} \\
\midrule
Correct
& 5
& 5/5 (100\%)
& 5/5 (100\%)
& 1.00 \\

Confusable
& 5
& 1/5 (20\%)
& 3/5 (60\%)
& 1.80 \\
\midrule
Overall
& 10
& 6/10 (60\%)
& 8/10 (80\%)
& 1.40 \\
\bottomrule
\end{tabular}%
}
\end{table}

\begin{figure*}[t]
    \centering
    \includegraphics[
        width=\textwidth,
        height=0.18\textheight,
        keepaspectratio
    ]{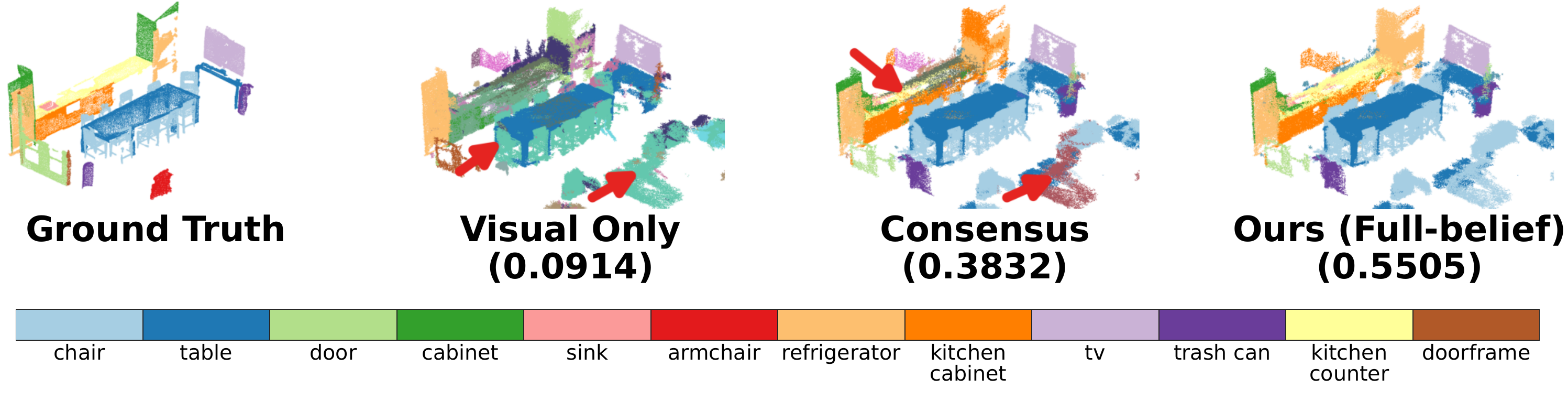}
    \caption{Qualitative effect of task-time semantic readout on a
    representative ScanNet200 scene. Parentheses report scene-level
    mIoU. Consensus recovers the semantic structure missed by visual-only
    prediction, while full-belief projection resolves additional class confusions.}
    \label{fig:ablation_semseg}
\end{figure*}

\subsection{Ablation Studies}
\label{sec:ablations}

We separately ablate semantic belief construction and task-time readout
while keeping object geometry and observations fixed.
Table~\ref{tab:component_ablation} isolates belief construction using the
consensus readout, whereas Table~\ref{tab:semseg_ablation} and
Fig.~\ref{fig:ablation_semseg} compares alternative semantic readouts.

Table~\ref{tab:component_ablation} shows that uniform evidence weighting
reduces both segmentation and navigation performance. Linear support
weighting slightly improves mIoU but reduces navigation success, particularly
on inserted YCB objects, indicating that observation size alone is an
imperfect proxy for semantic quality. Removing low-evidence regularization
accounts for the largest overall navigation drop. Removing synonym consolidation
causes a small decrease in mIoU and overall navigation success while leaving
YCB success unchanged.

Table~\ref{tab:semseg_ablation} shows that preserving additional semantic
hypotheses improves fixed-vocabulary projection. On ScanNet200, mIoU increases
from 0.1286 with visual-only prediction to 0.2107 with consensus, 0.2393 with
early commitment, and 0.2742 with full-belief projection. On Replica,
full-belief projection similarly obtains the highest mIoU, 0.2912.

Figure~\ref{fig:ablation_semseg} provides a qualitative example of this
progression. On the illustrated ScanNet200 scene, scene-level mIoU increases
from 0.0914 with visual-only prediction to 0.3832 with consensus and 0.5505
with full-belief projection.

\definecolor{fullrow}{RGB}{210, 235, 245}      
\definecolor{bestcell}{RGB}{186, 230, 201}     
\definecolor{secondcell}{RGB}{255, 224, 178}   

\begin{table}[t]
\centering
\caption{Belief-construction ablation using the consensus semantic
readout. Navigation
uses 78 HM3D--YCB trials, including 18 inserted-object trials.}
\label{tab:component_ablation}
\small
\setlength{\tabcolsep}{3.2pt}
\resizebox{\columnwidth}{!}{%
\begin{tabular}{lccc}
\toprule
Variant
& ScanNet200 mIoU $\uparrow$
& Nav SR@1m $\uparrow$
& YCB SR@1m $\uparrow$ \\
\midrule
\rowcolor{fullrow}
Default belief construction
& \second{0.2107}
& \best{76.9} (60/78)
& \best{61.1} (11/18) \\
$w_e=1$
& 0.2010
& 70.5 (55/78)
& \second{55.6} (10/18) \\
$w_e=c_en_e$
& \best{0.2117}
& 74.4 (58/78)
& 50.0 (9/18) \\
No synonym consolidation
& 0.2067
& \second{75.6} (59/78)
& \best{61.1} (11/18) \\
No low-evidence regularization
& 0.2095
& 70.5 (55/78)
& 50.0 (9/18) \\
\bottomrule
\end{tabular}}
\vspace{1mm}

\end{table}

\begin{table}[t]
\centering
\caption{Semantic readout ablation. Higher mIoU is better.
Best results per dataset are highlighted in green (bold).}
\label{tab:semseg_ablation}
\setlength{\tabcolsep}{5.5pt}
\begin{tabular}{llc}
\toprule
Dataset & Semantic readout & mIoU $\uparrow$ \\
\midrule
\multirow{4}{*}{ScanNet200}
& Visual-only & 0.1286 \\
& Consensus-only & 0.2107 \\
& Early-commit & 0.2393 \\
& Full-belief projection & \best{0.2742} \\
\midrule
\multirow{4}{*}{Replica}
& Visual-only & 0.2287 \\
& Consensus-only & 0.2831 \\
& Early-commit & 0.2701 \\
& Full-belief projection & \best{0.2912} \\
\bottomrule
\end{tabular}
\end{table}
\section{Conclusion}

We introduced \ours{}, an evidence-preserving object memory for
open-vocabulary language-guided navigation. By retaining observation-level
linguistic evidence, reliability cues, and frame--mask provenance alongside
aggregate geometry and a separate aggregate visual feature, \ours{} defers
semantic commitment to task-time readouts. Over the same frozen object graph,
full-belief fixed-vocabulary projection achieves the highest mIoU among the
evaluated readouts on ScanNet200 and Replica, while consensus and
early-commit retrieval obtain the highest navigation success under the
HM3D--YCB protocol. In 10 matched real-world cases comprising 20 Unitree G1
runs, execution-grounded verification and query-specific candidate exclusion
increase the observed intended-target confirmation rate from 6/10 to 8/10
without modifying the persistent graph or rewriting the instruction. These
results provide initial evidence that delayed semantic commitment allows
different downstream tasks to consume the same stored object evidence and
enables physical execution to correct an erroneous query-to-node binding when
the intended target remains represented in the map. The current evaluation is
limited to the reported datasets and scenes and to a pilot-scale real-world
study. Future
work will evaluate the framework across more diverse environments and extend
evidence-preserving memory to non-stationary scenes.







\bibliographystyle{ieeetr}   
\bibliography{references}

\end{document}